\pdfoutput=1
\documentclass[11pt]{article}

\usepackage{times}
\usepackage{latexsym}

\usepackage[T1]{fontenc}
\usepackage[utf8]{inputenc}

\usepackage{microtype}

\usepackage{enumitem}
\setlist{nolistsep}

\usepackage{booktabs}
\usepackage{graphicx}
\usepackage{comment}

\usepackage[T2A,T1]{fontenc}    
\usepackage[utf8]{inputenc}     
\usepackage[arabic,english]{babel}
\usepackage{CJKutf8}
\usepackage{soulutf8}
\iftrue % set to iftrue for camera ready version, iffalse to display draft comments
 \usepackage[final]{emnlp2021}
\usepackage[final]{proofing}

  \renewcommand\hl[1]{{#1}}  %% to remove the highlith
   {\draftnote{\red{#2}}}
   \newcommand\redHL[1]{}
  \newcommand\todo[1]{}
\newcommand{\Djame}[1]{}
\newcommand{\BM}[1]{}
\newcommand{\BS}[1]{}
\newcommand{\Antonis}[1]{}

\newcommand{\an}[1]{}
\newcommand{\bm}[1]{}

\else  % the stuff below is for the draft mode (hl, comments and so)
\usepackage[review]{emnlp2021}
\usepackage[draft]{proofing}

\newcommand{\Djame}[1]{
\textbf{\textcolor{red}{\hl{Djame: #1}}}
}

\newcommand{\BM}[1]{
\textbf{\textcolor{black}{\textcolor{green}{BM: #1}}}
}

\newcommand{\BS}[1]{
\textbf{\textcolor{blue}{\hl{BS: #1}}}
}

\newcommand{\Antonis}[1]{
\textbf{\textcolor{blue}{\hl{AA: #1}}}
}

\newcommand{\an}[1]{\textcolor{magenta}{[#1 --AA]}}
\newcommand{\bm}[1]{\textcolor{red}{[#1 --BM]}}

\newcommand\red[1]{{\textbf{\textcolor{red}{#1}}}}

\let\oldred\red
\renewcommand\red[1]{{\bf \oldred{{#1}}}}

 \newcommand\redHL[1]{\red{\hl{#1}}}
\let\olddraftnote\draftnote
\renewcommand\draftnote[1]{\olddraftnote{\red{#1}}}

\fi

\title{Synthetic Data Augmentation for Zero-Shot Cross-Lingual Question Answering}

\author{Arij Riabi$^{\ddagger}$\thanks{$^{\ast}$: equal contribution. The work of Arij Riabi was partly carried out while she was working at reciTAL.} \quad Thomas Scialom$^{\star\diamond\ast}$ \quad Rachel Keraron$^{\star}$\\{\bf Benoît Sagot$^{\ddagger}$ \quad Djamé Seddah$^{\ddagger}$ \quad Jacopo Staiano$^{\star}$} \\
$^\ddagger$ Inria, Paris, France\\
$^\diamond$ Sorbonne Universit\'e, CNRS, LIP6, F-75005 Paris, France\\
$^\star$ reciTAL, Paris, France \\
  {\tt \{thomas,rachel,jacopo\}@recital.ai} \\
  {\tt \{arij.riabi,benoit.sagot,djame.seddah\}@inria.fr} \\}

\begin{document}
\begin{CJK*}{UTF8}{gbsn}

\maketitle
\begin{abstract}
Coupled with the availability of large scale datasets, deep learning architectures have enabled rapid progress on \draftreplace{the Question Answering task}{Question Answering tasks}. However, most of those datasets are in English, and the performances of state-of-the-art multilingual models are significantly lower when evaluated on non-English data.
Due to high data collection costs, it is not realistic to obtain annotated data for each language one desires to support. 

We propose a method to improve \draftremove{the} Cross-lingual Question Answering performance without requiring additional annotated data, leveraging Question Generation models to produce synthetic samples in a cross-lingual fashion.
We show that the proposed method allows to significantly outperform the baselines trained on English data only\draftreplace{. We report}{, establishing thus} a new state-of-the-art on four \draftadd{multilingual} datasets: MLQA, XQuAD, SQuAD-it and PIAF (fr). 
% Performance on Question Answering has rapidly progress with deep learning architectures conjugated with large scale datasets. 
% However all these datasets are in English, and results don't meet expectations when state-of-the-art multilingual models are evaluated in Non-English data. 
% we rely on Question Generation to synthetic data in a cross-lingual fashion
%In a Question Answering task, the input is typically composed of 2 text sequences: i) the context, and ii) the question; we consider both scenarios: i) when both the question and the context belong to the same language (\emph{same-lang}); and ii) when the language of the question can differ from the language of the context (\emph{cross-lang}), allowing richer interactions such as querying a set of multilingual documents. %with a relative improvement in terms of Exact Match of more than $60\%$. 

%Most of recent works lie into self-supervised models trained in a multilingual fashion, then fine-tuned on the English downstream task. 
%First attempt to improve multilingual QA beyond zero shot from English data. 

\end{abstract}

\section{Introduction}

Question Answering is a fast-growing research field, aiming to improve the capabilities of machines to read and understand documents. Significant progress has recently been enabled by the use of large pre-trained language models \cite{devlin2018bert,raffel2019exploring}, which reach human-level performances on several publicly available benchmarks, such as SQuAD \cite{rajpurkar2016squad} and NewsQA \cite{trischler2016newsqa}.

Given that the majority of large scale Question Answering (QA) datasets are in English \cite{hermann2015teaching, rajpurkar2016squad, choi2018quac}, the development of QA systems targeting other languages is currently addressed via two cross-lingual QA datasets: XQuAD \cite{artetxe2019cross} and MLQA \cite{lewis-etal-2020-mlqa}, covering respectively 10 and 7 languages. Due to the cost of annotation, both are limited only to an evaluation set. They are comparable to the validation set of the original SQuAD (see more details in Section~\ref{subsec:eval_sets}). In both datasets, each paragraph is paired with questions in various languages, allowing to evaluate models in a \emph{cross-lingual} experimental scenario: the input context and the question can be in two different languages. This scenario has important practical applications, such as querying a set of documents in various languages. 

Performing this cross-lingual task is complex and remains challenging for current models, assuming only English training data: transfer results are shown to rank behind training-language performance \cite{artetxe2019cross, lewis-etal-2020-mlqa}. In other words, multilingual models fine-tuned only on English data are found to perform significantly better on English than on other languages. 
%To the best of our knowledge, no alternatives methods to such a simple zero-shot transfer have been proposed so far. 
Besides the almost simultaneous work of \citet{shakeri2020multilingual}, very few alternatives to such a simple zero-shot transfer method have been proposed so far.

In this paper, we propose to generate synthetic data in a cross-lingual fashion, borrowing the idea from monolingual QA research efforts \cite{duan2017question}. On English corpora, generating synthetic questions has shown to significantly improve the performance of QA models \cite{du-etal-2017-learning, golub2017two, du-cardie-2018-harvesting, alberti2019synthetic}.
However, the adaptation of this technique to cross-lingual QA is not straightforward: cross-lingual text generation is a challenging task \emph{per se} which has not been yet extensively explored, in particular when no multilingual training data is available. 

We explore two Question Generation scenarios: (i) requiring only SQuAD data; and (ii) using a translator tool to obtain translated versions of SQuAD. %Leveraging on such synthetic cross-lingual data, we also propose a training method better suited to cross-lingual tasks.
As expected, the method leveraging on a translator has shown to perform the best. Leveraging on such synthetic data, our best model obtains significant improvements on XQuAD and MLQA over the state-of-the-art for both Exact Match and F1 scores.
In addition, we evaluate the QA models on languages not seen during training (even for the synthetic data) -- using SQuAD-it (for Italian), PIAF (for French), and KorQUaD (for Korean) -- reporting a new state-of-the-art for Italian and French, and observing significant improvements on Korean compared to zero-shot without augmentation.
% on all considered datasets. %relative improvement of $18\%$ w.r.t. the baseline.
This indicates that the proposed method allows to capture better multilingual representations beyond the training languages. 
Our method paves the way toward multilingual QA domain adaptation, especially for under-resourced languages.

% \draftadd{
Our contributions can be summarized as follows:
\begin{itemize}
    \item We present a data augmentation approach for Cross-Lingual Question Answering based on synthetic Question Generation;
    \item We report extensive experiments showing significant improvements on two multilingual evaluation datasets (XQuAD and MLQA);
    \item We additionally evaluate the proposed methodology on languages unseen during training, thus showing the potential benefits for QA on low-resource languages.
\end{itemize}
% }

\section{Related Work}

\paragraph{Question Answering (QA)}
QA is the task for which given a context and a question, a model has to find the answer. The interest for Question Answering goes back a long way: in a 1965 survey, 
\citet{simmons1965answering} reported fifteen implemented English language question-answering
systems.
More recently, with the rise of large scale datasets \cite{hermann2015teaching}, and large pre-trained models \cite{devlin2018bert}, the performance drastically increased, approaching human-level performance on standard benchmarks -- see for instance the SQuAD leader board.\footnote{\url{https://rajpurkar.github.io/SQuAD-explorer/}}
More challenging evaluation benchmarks have recently been proposed: \citet{dua2019drop} released the DROP dataset, for which the annotators were encouraged to provide adversarial questions; \citet{Burchell2020Querent} released the MSQ dataset, consisting of multi-sentence questions. % for which the model needs to reason. 

However, all these works are focused on English. Another popular research direction focuses on the development of multilingual QA models. For this purpose, the first step has been to provide the community with multilingual evaluation sets: \citet{artetxe2019cross} and \citet{lewis-etal-2020-mlqa} concurrently proposed two different evaluation sets which are comparable to the SQuAD development set. Both reach the same conclusion: due to the lack of non-English training data, models do not achieve the same performance in Non-English languages than they do in English.
To the best of our knowledge, no method has been proposed to fill this gap.

\paragraph{Question Generation (QG)}
QG can be seen as the dual task of QA: the input is composed of the \emph{answer} and the \emph{paragraph} containing it, and the model is trained to generate the \emph{question}. Proposed by \citet{rus2010first}, it has leveraged on the development of new QA datasets \cite{zhou2017neural, scialom2019self}. Similar to QA, significant performance improvements have been obtained using pre-trained language models \cite{dong2019unified}. Still, due to the lack of multilingual datasets, most previous works have been limited to monolingual text generation. We note the exceptions of \citet{kumar2019cross} and \citet{chi2020cross}, who resorted to multilingual pre-training before fine-tuning on monolingual downstream NLG tasks. However, the quality of the generated questions is still found inferior to the corresponding English ones.

\paragraph{Question Generation for Question Answering}
Data augmentation via synthetic data generation is a well-known technique to improve models' accuracy and generalisation. It has found successful application in several areas, such as time series analysis \cite{forestier2017generating} and computer vision \cite{buslaev2020albumentations}.
In the context of QA, generating synthetic questions to complete a dataset has shown to improve QA performances \cite{duan2017question, alberti2019synthetic}. So far, all these works have focused on English QA given the difficulty to generate questions in other languages without available data. 
This lack of data, and the difficulty to obtain some, constitutes the main motivation of our work and justifies exploring cost-effective approaches such as data augmentation via the generation of questions.

%In a very recent work, almost simultaneous to ours, \citet{shakeri2020multilingual} address multilingual QA with a similar approach. We detail the differences in our discussion, Section~\ref{discussion}.
In a very recent work, almost simultaneous to our previously submitted version, \citet{shakeri2020multilingual} address multilingual QA with a similar approach. However, we argue that their experimental protocol does not allow to totally answer the research question. We detail the differences in our discussion, Section~\ref{discussion}.

\section{Data}

\subsection{English Training Data}

\paragraph{SQuAD$_{en}$} The original SQuAD \cite{rajpurkar2016squad}, which we refer as SQuAD$_{en}$ for clarity in this paper. It is one of the first, and among the most popular, large scale QA datasets. It contains about 100K question/paragraph/answer triplets in English, annotated via Mechanical Turk.\footnote{Two versions of SQuAD have been released: v1.1, used in this work, and v2.0. The latter contains ``unanswerable questions'' in addition to those from v1.1. We use the former, since the multilingual evaluation datasets, MLQA and XQUAD, do not include unanswerable questions.}

\paragraph{QG datasets}
Any QA dataset can be reversed into a QG dataset, by switching the generation targets from the answers to the questions. In this paper, we use the \emph{qg} subscript to specify when the dataset is used for QG (e.g. SQuAD$_{en;qg}$ indicates the English SQuAD data in QG format).

\subsection{Synthetic Training Sets}

\paragraph{SQuAD$_{trans}$} is a machine translated version of the SQuAD train set in the seven languages of MLQA, released by the authors together with their paper. 

\paragraph{WikiScrap} 
We collected 500 Wikipedia articles for all the languages present in MLQA. 
They are not paired with any question or answer. We use them as contexts to generate synthetic multilingual questions, as detailed in Section~\ref{sec:SyntheticDataAugmentation}. Following the SQuAD$_{en}$ protocol, we used project Nayuki's code\footnote{\url{https://www.nayuki.io/page/computing-wikipedias-internal-pageranks}} to parse the top 10K Wikipedia pages according to the PageRank algorithm \cite{page1999pagerank}. We then filtered out paragraphs with character length outside of a [500, 1500] interval. Articles with less than 5 paragraphs are discarded, since they tend to be less developed, in a lower quality or being only redirection pages. Out of the filtered articles, we randomly selected 500 per language.

\subsection{Multilingual Evaluation Sets}
\label{subsec:eval_sets}

\paragraph{XQuAD} \cite{artetxe2019cross} is a human translation of the SQuAD$_{en}$ development set in 10 languages (Arabic, Chinese, German, Greek, Hindi, Russian, Spanish, Thai, Turkish, and Vietnamese), providing 1k QA pairs for each language.

\paragraph{MLQA} \cite{lewis-etal-2020-mlqa} is an evaluation dataset in 7 languages (English, Arabic, Chinese, German, Hindi, and Spanish). The dataset is built from aligned Wikipedia sentences across at least two languages (full alignment between all languages being impossible), with the goal of providing natural rather than translated paragraphs. The QA pairs are manually annotated on the English sentences and then human translated on the aligned sentences. The dataset contains about 46k aligned QA pairs in total. %The authors also provide a machine translation in the seven languages of the SQuAD dev and train data, as well as a machine translation to English of the non-English MLQA data.

\paragraph{Language-specific benchmarks}
In addition to the two aforementioned multilingual evaluation corpora, we benchmark our models on three language-specific datasets for French, Italian and Korean, as detailed below. 
We choose these datasets since none of these languages are present in XQuAD or MLQA. Hence, they allow us to evaluate our models in a scenario where the target language is not available during training, even for the synthetic questions.

\paragraph{PIAF} \newcite{keraron2020project} provided an evaluation set in French following the SQuAD protocol, containing 3835 examples. 

\paragraph{KorQuAD 1.0} the Korean Question Answering
Dataset \cite{lim2019korquad1}, a Korean dataset also built following the SQuAD protocol. 

\paragraph{SQuAD-it} Derived from SQuAD$_{en}$, it was obtained via semi-automatic translation to Italian \cite{10.1007/978-3-030-03840-3_29}.

\section{Models}
\label{sec:models}
Recent works \cite{raffel2019exploring, lewis2019bart} have shown that classification tasks can be framed as a text-to-text problem, achieving state-of-the-art results on established benchmarks, such as GLUE \cite{wang2018glue}.
Accordingly, we employ the same architecture for both Question Answering and Generation tasks.
This also allows fairer comparisons for our purposes, by removing differences between QA and QG architectures and their potential impact on the results obtained.
In particular, we use a distilled version of XLM-R \cite{conneau2019unsupervised}:  MiniLM-M \cite{wang2020minilm} (see Section~\ref{subsec:implem_detail} for further details). 

\subsection{Baselines}

\paragraph{QA\textsubscript{No-synth}} Following previous works, we fine-tuned the multilingual models on SQuAD$_{en}$, and consider them as our baselines.

\paragraph{English as Pivot} Leveraging on translation models, we consider a second baseline method, which uses English as a pivot. First, both the question in language $L_{q}$ and the paragraph in language $L_{p}$ are translated into English.
We then invoke the baseline model described above, QA\textsubscript{No-synth}, to predict the answer. Finally, the predicted answer is translated back into the target language $L_{p}$.
%In theory, given a perfect translation, the scores should reach a similar performance than those on SQuAD$_{en}$ test. In practice, we observe poor performances compared to the baseline. As an additional downside, this method requires the translation to be performed not only at training time, but also during inference. 
We used the google translate API.\footnote{\url{https://translate.google.com}} 

\paragraph{QA\textsubscript{+SQuAD-trans}} the translated data SQuAD\textsubscript{trans} are used as additional training data to SQuAD\textsubscript{en}, to train the QA model.

\subsection{Question Generation Data Augmentation}
\label{sec:SyntheticDataAugmentation}

\begin{comment}
\begin{table*}
\centering\small
\begin{tabular}{llllr}
\toprule

Data                    & QG Model          & QG Train Data      & Synth. Data Used       & \# Synth.     \\\midrule
%$No-synth$ (Baseline)   & N/A               & SQuAD$_{en}$    & N/A                       & 0             \\ \hline
\emph{No-trans}              & QG\textsubscript{No-trans}   & SQuAD\textsubscript{{en}}    & WikiScrap                 & 106K           \\
%\emph{Same-lang}             & QG\textsubscript{Same-lang}  & SQuAD\textsubscript{trans} & WikiScrap+SQuAD\textsubscript{trans} & 416K           \\
\emph{Cross-lang}            & QG\textsubscript{Cross-lang} & SQuAD\textsubscript{trans} & WikiScrap+SQuAD\textsubscript{trans} & 416K           \\
\bottomrule
\end{tabular}%
\caption{Summary of the different setups for our QA models, given different synthetic data.}
\label{tab:summar_qa_differences}
\end{table*}

\end{comment}

In this work we consider data augmentation via generating synthetic questions, to improve the QA performance. Different training schemes for the question generator are possible, resulting in different quality of the synthetic data. Before this work, its impact on the final QA system remained unexplored in a multilingual context. 

For all the following experiments, only the synthetic data changes. Given a specific set of synthetic data, we always follow the same two-stages protocol, similar to \citet{alberti2019synthetic}: we first train the QA model on the synthetic QA data, then on SQuAD$_{en}$. 
We also tried to train the QA model in one stage, with all the synthetic and human data shuffled together, but observed no improvements over the baseline. 

We explored two different synthetic generation modes: % and summarized in Table~\ref{tab:summar_qa_differences}.

\paragraph{Synth}
the QG model is trained on SQuAD$_{en,qg}$ (i.e., English data only) and the synthetic data are generated on WikiScrap. Under this setup, the only annotated samples this model has access to are those from SQuAD-en.

%\paragraph{Same-lang}: The QG model is trained on SQuAD$_{trans,qg}$ in addition to SQuAD$_{en,qg}$. The context input and the question to generate, both have to be in the same language $X \in SQuAD_{trans} languages$ (e.g. given an Arabic context, the model is trained to generate an Arabic question). The synthetic data are generated on WikiScrap and SQuAD$_{trans}$.

\paragraph{Synth+trans}
the QG model is trained on SQuAD$_{trans,qg}$ in addition to SQuAD$_{en,qg}$. 
The questions can thus be in a different languages than the context. Hence, the model needs an indication about the language it is expected to generate the question in. 
To control the target language, we use a specific prompt per language, defining a special token \texttt{<LANG>}, which corresponds to the desired target language $Y$. Thus, the input is structured as \texttt{<LANG> <SEP> Answer <SEP> Context},
% $<LANG>$ \space $<SEP>$  \space $Context$
where \texttt{<LANG>} indicates to the model in what language the question should be generated, and \texttt{<SEP>} is a special token acting as a separator. These attributes offer flexibility on the target language. Similar techniques are used in the literature to control the style of the output \cite{keskar2019ctrl, scialom2020toward, chi2020cross}.

\subsection{Implementation details}
\label{subsec:implem_detail}
For all our experiments we use Multilingual MiniLM v1 (MiniLM-m) \cite{wang2020minilm}, a 12-layer with 384 hidden size architecture distilled from XLM-R Base multilingual \cite{conneau2019unsupervised}. With only 66M parameters, it is an order of magnitude smaller than state-of-the-art architectures such as BERT-large or XLM-large.
We used the official Microsoft implementation.\footnote{Publicly available at \url{https://github.com/microsoft/unilm/tree/master/minilm}.} For all the experiments --both QG and QA-- we trained the model for 5 epochs, using the default hyper-parameters. We used a single nVidia gtx2080ti with 11G RAM, and the training times
% \draftreplace{amounts}{amount} 
amount to circa 4 and 2 hours for Question Generation and for Question Answering, respectively. 
To evaluate our models, we used the official MLQA evaluation scripts.\footnote{\url{https://github.com/facebookresearch/MLQA/blob/master/mlqa_evaluation_v1.py}}
For reproducibility purposes, we make the code available.\footnote{\url{https://anonymous.4open.science}}

\section{Results}

\subsection{Question Generation}
\label{sec:res_QG}
We report examples of generated questions in Table~\ref{tab:examples_generated_questions}.
%https://docs.google.com/document/d/19tGtTsS2wGH0ucKr5_Zxk4NSA61Tn8T4E7dI-id0WcA/edit

\begin{table*}
    \centering\small
    \begin{tabular}{p{.95\linewidth}}
    \toprule
    
        \textbf{Paragraph (EN)}
       Peyton Manning became the first quarterback ever to lead two different teams to multiple Super Bowls. He is also the oldest quarterback ever to play in a Super Bowl at age 39. The past record was held by John Elway, who led the Broncos to victory in Super Bowl XXXIII at age 38 and is currently Denver's Executive Vice President of Football Operations and General Manager.
       
        \textbf{Answer} 
        Broncos
        
        \textbf{QG\textsubscript{synth} }
        What team did John Elway lead to victory at age 38?
        
        %\textbf{QG\textsubscript{Same-lang}}
        % What team did John Elway lead?

        \textbf{QG\textsubscript{synth+trans} (target language = en)}
        What team did John Elway lead to win in the Super Bowl?
        \\
        \midrule
        
        \textbf{Paragraph (ES)}
        Peyton Manning se convirtió en el primer mariscal de campo de la historia en llevar a dos equipos diferentes a participar en múltiples Super Bowls. Ademas, es con 39 años, el mariscal de campo más longevo de la historia en jugar ese partido. El récord anterior estaba en manos de John Elway ―mánager general y actual vicepresidente ejecutivo para operaciones futbolísticas de Denver― que condujo a los Broncos a la victoria en la Super Bowl XXXIII a los 38 años de edad.

        \textbf{Answer} 
        Broncos
        
        \textbf{QG\textsubscript{synth} }
        Where did Peyton Manning condujo?
        
        %\textbf{QG\textsubscript{Same-lang}}
        %Qué equipo ganó la super Bowl de los 38 años de edad? (\emph{Which team won the 38-year-old Super Bowl?})
        
        \textbf{QG\textsubscript{synth+trans} (target language = es)}
        Qué equipo ganó el récord anterior? (\emph{Which team won the previous record?})
        
        \textbf{QG\textsubscript{synth+trans} (target language = en)}
        What team did Menning win in the Super Bowl?
        \\
        \midrule

        \textbf{Paragraph (ZH)}
       培顿·曼宁成为史上首位带领两支不同球队多次进入超级碗的四分卫。他也以 39 岁高龄参加超级碗而成为史上年龄最大的四分卫。过去的记录是由约翰·埃尔维保持的，他在 38岁时带领野马队赢得第 33 届超级碗，目前担任丹佛的橄榄球运营执行副总裁兼总经理
       
        \textbf{Answer} 
        野马队
        
        \textbf{QG\textsubscript{synth}}
        What is the name for the name that the name is used?
        
        %\textbf{QG\textsubscript{Same-lang}}
        %约翰·埃尔维赢得超级碗是什么队? (\emph{What team did John Elvey win the Super Bowl?})
        
        \textbf{QG\textsubscript{synth+trans} (target language = zh)}
         约翰·埃尔维在13岁时带领哪支球队赢得第 33届超级碗? (\emph{Which team did John Elvey lead to win the 33rd Super Bowl at the age of 13?})
         
        \textbf{QG\textsubscript{synth+trans} (target language = en)}
        What team won the 33th Super Bowl?
        \\
        \bottomrule
    \end{tabular}
    \caption{Example of questions generated by the different models on an XQuAD's paragraph in different languages. For QG\textsubscript{synth+trans}, we report the outputs given two target languages, the one of the context and English.}
    \label{tab:examples_generated_questions}
\end{table*}

\paragraph{Controlling the Target Language}

In the context of multilingual text generation, controlling the target language is not trivial. 

When a QA model is trained only on English data, at inference, given a non-English paragraph, it predicts the answer in the input language, as one would expect, since it is an extractive process.
Ideally, we would like to observe the same behavior for a Question Generation model trained only on English data (such as $Synth$), leveraging on the multilingual pre-training. Conversely to QA, QG is a language generation task. \emph{Multilingual} generation is much more challenging, as the model's decoding ability plays a major role. 
When a QG model is fine-tuned only on English data (i.e SQuAD-en), its controllability of the target language suffers from catastrophic forgetting: the input language does not propagate to the generated text. While still relevant to the context, the synthetic questions are generated in English: for instance, in Table~\ref{tab:examples_generated_questions} we observe that the QG\textsubscript{synth} model outputs English questions for the paragraphs in Chinese and Spanish. The same phenomenon was reported by \citet{chi2020cross}. 

%\paragraph{Same language training}
%To tackle the aforementioned limitation on target language controllability (i.e. to enable the generation in other languages than English), one needs multilingual data. We can leverage on the translated version of the dataset to add the needed non-English examples, corresponding to $QG_{No−trans}$ (see Table~\ref{tab:summar_qa_differences}. in Table~\ref{tab:examples_generated_questions}, we observe that for $QG_{Same−lang}$, the questions are generated in the same language than the input. It indicates that the model succeed in this aspect of the decoding. Nonetheless, the questions are more relevant and coherent compared to $QG_{No−trans}$: for the Spanish paragraph, the question focus well on the input answer \emph{Broncos} and is well formed. It contains only a small mistake about the \emph{38 years old}, that is the age of the captain John Elway instead of the Super Bowl age. Still, this error could be due to an encoding co-reference problem, while the decoding stage seems to work efficiently that way. 

\paragraph{Cross-Lingual Training}
To overcome the aforementioned limitation on target language controllability (i.e. to enable the generation in other languages than English), multilingual data is needed. We can leverage on the translated versions of the dataset to add the required non-English examples. 
As detailed in Section~\ref{sec:SyntheticDataAugmentation}, we simply use a specific prompt that corresponds to the target language (with $N$ different prompts corresponding to the $N$ languages present in the dataset). 
In Table~\ref{tab:examples_generated_questions}, we show how QG\textsubscript{synth+trans} can generate questions in the same language as the input. 
% This indicates that the generator succeeds in this aspect of the decoding. 
These synthetic questions seem much more relevant, coherent and fluent, if compared to those produced by QG\textsubscript{synth}: for the Spanish paragraph, the question is well formed and focused on the input answer; for Chinese (see bottom row of Table~\ref{tab:examples_generated_questions} for QG\textsubscript{synth+trans}) is perfectly written. %Note that it refers to the \emph{33th} Super Bowl, while in the original input it was written using the Roman equivalent (\emph{XXXIII}): this comes from the translation in Chinese as you can see written 33 in Chinese.

\begin{table}[!h]
    \centering \small
    \begin{tabular}{llllllll} \hline
q/c & en   & es  & de  & ar  & hi  & vi   & zh   \\ \hline
en & 14.5  & 8.9 & 7.2 & 5.9 & 6.5 & 8.4  & 6.0   \\
es &  9.0  & 10. & 6.6 & 4.2 & 5.9 & 6.3  & 4.6  \\
de &  6.2  & 4.8 & 6.3 & 3.1 & 3.7 & 5.0  & 3.2  \\
ar &  2.8  & 2.2 & 2.4 & 3.3 & 2.0 & 2.3  & 2.1   \\
hi &  7.9  & 6.7 & 6.6 & 5.8 & 8.3 & 6.6  & 5.2  \\
vi &  9.1  & 7.3 & 7.2 & 6.0 & 6.5 & 12.3 & 6.1   \\
zh &  9.2  & 8.0 & 7.8 & 6.1 & 7.2 & 8.0  & 15.0  \\ \hline
\end{tabular}
    \caption{BLEU-4 scores on MLQA test for QG\textsubscript{synth+trans}. Columns show context language, rows show question language.}
    \label{tab:QG_MLQA_synth+trans}
\end{table}

In Table~\ref{tab:QG_MLQA_synth+trans} we report the BLEU4 scores for QG\textsubscript{synth+trans} grouped by the language of the question. 
As expected, the score is maximized on the diagonal (same languages for the context and the question). Still, most of these scores are lower on non-English languages. It is interesting to note BLEU4 correlates with the QA scores: 0.51 Pearson coefficient. The reasons are two folds: 1) QA and QG share the same Language Model, which might struggle for the same languages; 2) the better the QG, the better the synthetic data, therefore the better the QA performs. 
We discuss further in Section~\ref{discussion} how this impacts the QA performance. 

In addition to BLEU, we also report the QA F1 scores for different QA models when applied on the generated questions in the supplementary material. 
Yet, we warn the reader that these results should be taken with caution: evaluating NLG is known to be an open research problem; BLEU is known to suffer from important limitations \cite{novikova-etal:2017:need-new-metrics-nlg}, which might be accentuated in a multilingual context \cite{lee2020reference}. For this reason, we conducted a manual qualitative analysis on a small number of samples. Note that the annotators need to have a professional level in the language of the generated question to evaluate its fluency, and to be bilingual, when evaluating its relevance w.r.t.~input context in our cross language scenario. This is a significant challenge to conduct a large scale evaluation.

So far, our results (see at the end of Supplementary Material) for Arabic and German show an overall good quality in the questions: only one question for Arabic was genuinely missing the point while for German there were 2 lexical questionable choices that invalidate the question (out of 10 samples for both languages so far). This indicates that Arabic questions could actually be better than what their low BLEU score shows. Arabic has a very different morphological structure that could explain such low BLEU \cite{bouamor2014human}. This emphasizes the limitation of the current automatic metrics in a multilingual context. 

\subsection{Question Answering}

\begin{table*}
\centering
\begin{tabular}{lcccc}

                                          & \#Params & Trans. & XQuAD   & MLQA        \\ \midrule
MiniLM \cite{wang2020minilm}                          & 66M  & No  & 42.2 / 29.5          & 38.4 / 26.0   \\
XLM \cite{hu2020xtreme}\footnote{We used the script released by the authors, see \url{https://github.com/google-research/xtreme/blob/master/scripts/train_qa.sh}}           & 340M     & No     &  68.5/52.8          & \textbf{65.4} / 47.9      \\
\midrule

English as Pivot                                     & 66M  & Yes  & 46.2 / 30.9          & 36.1 / 23.0   \\
$MiniLM_{+synth}$                                    & 66M  & No   & 44.8 / 33.1          & 39.8 / 27.5   \\
$MiniLM_{+SQuAD-trans}$                              & 66M  & Yes  & 55.0 / 40.7          & 49.5 / 35.3   \\
$MiniLM_{+synth-trans}$                              & 66M  & Yes  & 63.3 / 49.1          & 56.1 / 41.4   \\
$MiniLM_{+SQuAD-trans+synth-trans}$                  & 66M  & Yes  & 62.5 / 48.6          & 55.0 / 40.4   \\
$XLM-R_{+synth-trans}$                                 & 340M  & Yes & \textbf{74.3 / 59.2} & \textbf{65.3 / 49.2}  \\

\bottomrule
\end{tabular}

\caption{Results (F1 / EM) of the different QA models on XQuAD and MLQA. XLM corresponds to the large version.}
\label{tab:main_res}
\end{table*}

We report the main results of our experiments on XQuAD and MLQA in Table~\ref{tab:main_res}. The scores correspond to the average over all the different possible combination of languages (\emph{de-de}, \emph{de-ar}, etc.).  

\paragraph{English as Pivot}
Using English as a pivot does not lead to good results. This may be due to %the poor quality of the multiple translations for some languages, as well as to 
the evaluation metrics, which are based on $n$-grams similarity. For extractive QA, F1 and EM metrics measure the overlap between the predicted answer and the ground truth. Therefore, meaningful answers worded differently are penalized, a situation that is likely to occur because of the back-translation mechanism. This makes automatic evaluation challenging for this setup, as metrics suffer from similar difficulties as those observed for text generation \cite{sulem2018bleu}.
%However, the scores are consistently lower compared to the baseline. 
As an additional downside, this model requires multiple translations \emph{at inference time}. For these reasons, we decided not to explore this approach further. 

\paragraph{Synthetic without translation (+synth)}
%compared to QA\textsubscript{-synth}, we observe a small performance decrease for the $Same-lang$ setup ($Same$ score in the Table~\ref{tab:main_res}). It can be explained by the low quality of the synthetic questions of $QG_{no-trans}$ (see Table~\ref{tab:examples_generated_questions} and Section~\ref{sec:res_QG}). 
Compared to the MiniLM baseline, we observe a small performance increase for MiniLM\textsubscript{+synth} (Exact Match increases from 29.5 to 33.1 on XQuAD and from 26.0 to 27.5 on MLQA). 

During the self-supervised pre-training stage, the model was exposed to multilingual inputs. Yet, for a given input, the target language was always consistent, preventing the model to be exposed to such a cross-lingual scenario.
The synthetic inputs are composed of questions
%mostly
in English (see examples in Table~\ref{tab:examples_generated_questions}) while the contexts can be in any languages. Therefore, the QA model is exposed for the first time to a cross-lingual scenario.
We hypothesise that such a cross-lingual ability is not innate for a default multilingual model: exposing a model to this scenario allows to develop this ability and contributes to improve its performance.

%\paragraph{QA\textsubscript{Same-lang}}:
%We observe a significant improvement over the baseline for both Same and Cross scores (+4.3\% Same / 4.8\% Cross on XQuAD and +3\% Same / +6\% Cross on MLQA; see in Table~\ref{tab:main_res}), indicating a positive impact of the proposed method.  

\paragraph{Synthetic with translation (+synth-trans)}:
For MiniLM\textsubscript{+synth-trans}, we obtain a much larger improvement over its baselines, MiniLM, compared to MiniLM\textsubscript{+synth}, on both MLQA and XQuAD. Also, it outperforms MiniLM\textsubscript{+SQuAD−trans}, indicating the benefit of our proposed approach. 
%Conversely, for the Cross score, the gain is way larger, improving by more than 10\% on both XQuAD and MLQA, over QA\textsubscript{Same-lang}.
This supports the intuition developed in the previous paragraph: independently of the multilingual capacity of the model, a cross-lingual ability is developed when the two inputs components are not exclusively written in the same language. In Section~\ref{discussion}, we discuss this phenomenon more in depth. 

\begin{table*}
\centering
%\resizebox{\columnwidth}{!}{
\begin{tabular}{lrrrr}
                                      & \#Params     & PIAF (fr)   & KorQuAD      & SQuAD-it     \\ \toprule
MiniLM   \cite{wang2020minilm}        & 66M          & 58.9 / 34.3 & 53.3 / 40.5  & 72.0 / 57.7     \\
mBert \cite{devlin2018bert}          & 110M         & 64.4 / 42.5 &  -           & 74.1 / 62.5     \\ 
CamemBERT \cite{martin2019camembert}  & 340M         & 68.9 / -    & $N/A$        & $N/A$            \\ 
%English as Pivot                     & 66M          & 51.9/27.4 & -       & -             \\
\midrule
MiniLM\textsubscript{+synth}          & 66M          & 58.6 / 34.5 & 52.1 / 39.0   & 71.3 / 58.0      \\ 
%synth\_wiki par QG\_cross + squad en                             & 62.81/39.17 & -\\ \hline
%synth\_wiki par QG\_cross + squad en + mlqa\_traduit             & 62.81/38.06 & - \\ \hline
%QA\textsubscript{Same-lang}    & 66M         & 63.76/40.36  \\
MiniLM\textsubscript{+synth-trans}    & 66M         & 63.9 / 40.6 & 60.0 / 48.8   & 74.5 / 62.0    \\ 
XLM-R\textsubscript{+synth-trans}       & 340M        & \textbf{72.1 / 47.1} & \textbf{63.0 / 52.8}    & \textbf{80.4 / 67.6}
\\ 
\bottomrule
%synth\_wiki+mlqa\_synth par QG\_cross + squad en + mlqa\_traduit &  63.94/40.39  & - \\ \hline
\end{tabular}%

%}
\caption{Zero-shot results (F1 / EM) on PIAF, KorQuAD and SQuAD-it for our different QA models, compared to various baselines. For mBert on SQuAD-it, we report the score from \citet{10.1007/978-3-030-03840-3_29}. Note that CamemBERT is a French version of RoBERTa, an architecture widely outperforming BERT.}
\label{tab:result_unseen_lang}
\end{table*}

\subsection{Discussion}
\label{discussion}

\paragraph{Cross Lingual Generalisation}

\begin{figure*}[!ht]
\centering
\includegraphics[width = .45\textwidth]{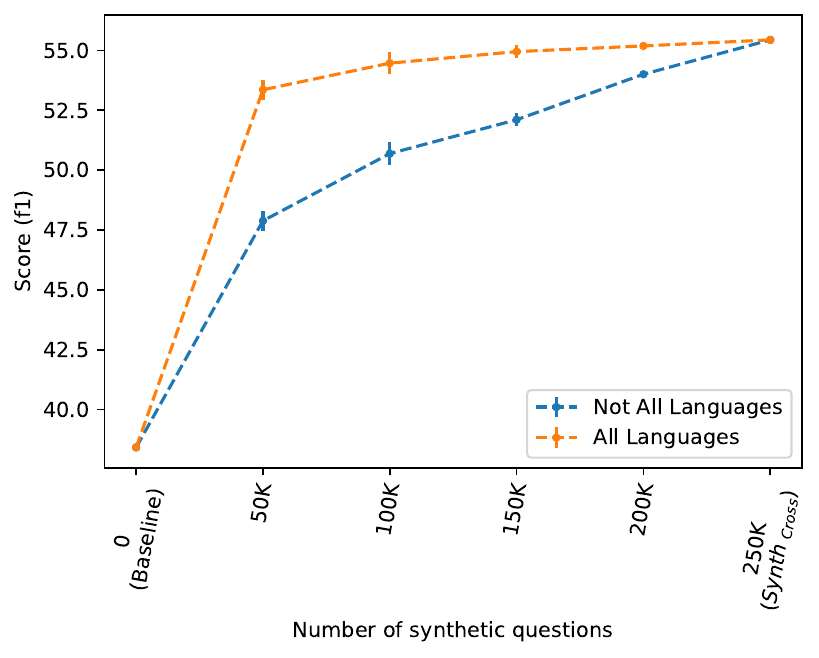}
\includegraphics[width = .45\textwidth]{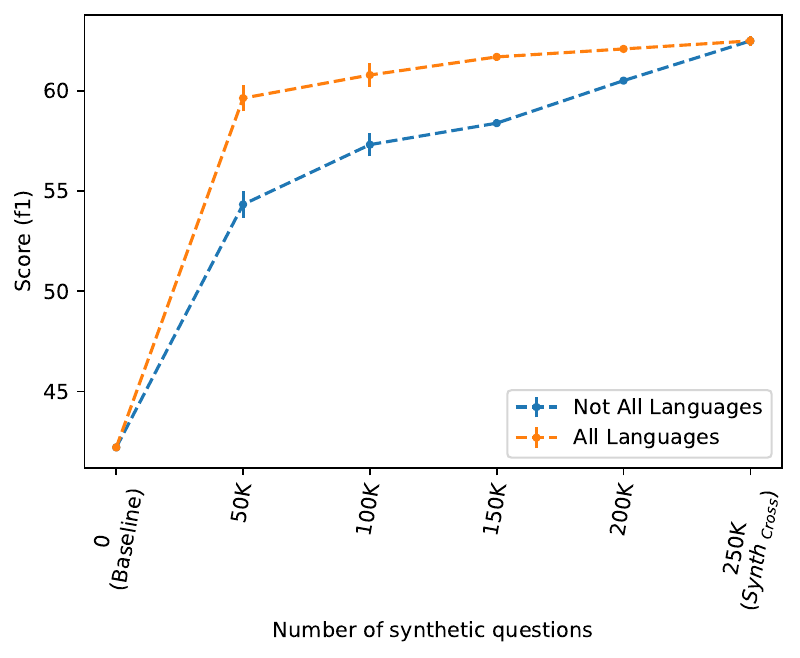}
\vspace{-0.2cm}
\caption{Left: F1 score on MLQA, for models with different number of synthetic data in two setups: for \emph{All Languages}, the synthetic questions are sampled among all the five languages in MLQA; for \emph{Not All Languages}, the synthetic questions are sampled progressively from only one language, two, \ldots, to all five for the last point, which corresponds to \emph{All Languages}. We report the standard deviation over five different permutations of the language ordering.
% We computed 5 seeds with different orders for the languages and also display the standard deviation. 
Note that, as expected, the more the synthetic data, the lower the variance in the results. 
Right: same as on the left, but evaluated on XQuAD. 
}
\vspace{-0.2cm}
\label{fig:cross_ability}
\end{figure*}

To explore the models' effectiveness in dealing with cross-lingual inputs, we report in Figure~\ref{fig:cross_ability} the performance for our MiniLM\textsubscript{+synth-trans} setup, varying the number of samples and the languages present in the synthetic data.
The abscissa $x$ corresponds to the progressively increasing number of synthetic samples used; at $x=0$, it corresponds to the MiniLM\textsubscript{+trans} baseline, where the model has access only to the original English data from SQuAD$_{en}$.
We explore two sampling strategies for the synthetic examples: 
\begin{enumerate}
    \item \emph{All Languages} corresponds to sampling the examples from any of the different languages. 
    \item  Conversely, for \emph{Not All Languages}, we progressively added the different languages: for $x=50K$, all the 50K synthetic data are on a unique language input, $L1$. Then for $x=100K$, the synthetic data are from either $L1$, or an additional language $L2$; finally, for $x=250K$, all MLQA languages are present. 
\end{enumerate}
 
In Figure~\ref{fig:cross_ability}, we observe that the performance for \emph{All Languages} increases largely at the beginning, then remains mostly stable. Conversely, we note a gradual improvement for \emph{Not All Languages}, as more languages are made available during training. 
This shows that when all the languages are present in the synthetic data, the model immediately develops cross-lingual abilities.

\begin{figure}[!ht]
\centering
\includegraphics[width = .45\textwidth]{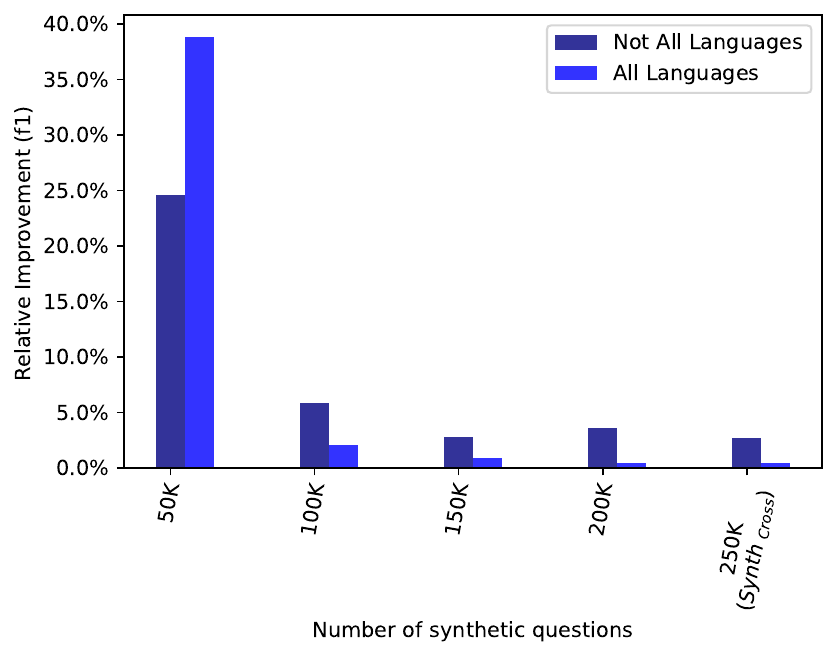}
\vspace{-0.2cm}
\caption{The relative variation in performance for the models in Figure~\ref{fig:cross_ability}.}
\label{fig:cross_ability_cum}
\vspace{-0.5cm}
\end{figure}

However, it appears that even with only one language pair present, the model is able to develop a cross-lingual ability that brings benefits on other languages: of Figure~\ref{fig:cross_ability_cum}, we can see that most of the improvement is happening given only one cross-lingual language pair (i.e. English and Spanish). 
% This could indicate that the model is able to generalise, in part, such ability, to other languages. 

\paragraph{Unseen Languages}

% PIAF

To measure the benefit of our approach on unseen languages (i.e. not present in the synthetic data from MLQA/XQuAD), we test our models on three QA evaluation sets:  PIAF (fr), KorQuAD  and SQuAD-it (see Section~\ref{subsec:eval_sets}). 
The results are consistent with the previous experiments on MLQA and XQuAD. 
Our MiniLM\textsubscript{+synth-trans} model outperforms its baseline by more than 4 Exact Match points, while XLM-R\textsubscript{+synth-trans} obtains a new state-of-the-art. Notably, our multilingual XLM-R\textsubscript{+synth-trans} outperforms CamemBERT on PIAF, even if the latter is a pure monolingual, in-domain language model.

%Note that since PIAF contexts are not aligned with the SQuAD-en ones. Hence, only the Same-lang score can be computed, not the Cross-lang one.

\paragraph{On the correlation between BLEU4 and QA scores}
To measure the impact of the quality of the generated questions on the QA performance, we computed the Pearson correlation between the BLEU4 and the QA scores. The coefficient is equal to 0.65 ($p<.001$). 
When we observe the correlations grouping the samples w.r.t.~their language question (i.e. the rows in Table~\ref{tab:QG_MLQA_synth+trans}), we obtain: \emph{en 0.94; es 0.84; de 0.46; ar 0.36; hi 0.33; vi 0.73; zh 0.92}. We observe stronger correlation for languages with higher BLEU scores (i.e en \& zh), and lower for the Arab that had the lowest BLEU, indicating an impact on the final QA score in par to the quality of the synthetic questions.

\paragraph{Differences with \citet{shakeri2020multilingual}}
A very recent
% \draftremove{, almost simultaneous,} 
work has addressed multilingual QA with a very similar approach. However, we note a major difference in our respective experiments regarding the choice for the QA and QG models. \citet{shakeri2020multilingual} choose  mBert for QA and T5-m for QG. We would like to emphasize that because T5-m significantly outperforms mBert it is not clear where the improvement comes from: is it due to the proposed approach, or simply from a distillation effect from T5-m to mBert? 
% \draftreplace{As opposed,}{In our case,}
In our case,
we deliberately used MiniLM \emph{for both QA and QG}: this allows a fairer investigation about the benefits of the proposed approach.

\begin{comment}

\paragraph{Differences with \citet{shakeri2020multilingual}}
A very recent, almost simultaneous, work has addressed multilingual QA with a similar approach. However, we note two major differences. 

First, the authors report the  for a specific subset of the possible language combinations: only when the paragraph and the question are written \emph{in the same language}. Instead, we report it for all the possible combinations. Our choice is motivated by \emph{1)} our goal of measuring the cross-lingual capabilities of the models, in their overall complexity; and \emph{2)}, to solve a practical use-case: it often happens that one asks a question in a language, while the answer is contained in document written in a different language. 

The second main difference is related to choice for the QA and QG models. \citet{shakeri2020multilingual} choose  mBert for QA and T5-m for QG. We would like to emphasize that because T5 significantly outperforms mBert it is not clear the improvement comes from the proposed approach, or simply from a distillation effect from T5-m to mBert. 
As opposed, we deliberately used MiniLM \emph{for both QA and QG}: this allows a fairer investigation about the benefits of our proposed approach.

The simultaneity and similarity of \citet{shakeri2020multilingual} and our work confirms the validity of the proposed approach and strengthens, if needed, the motivation to provide methods able to cope with the lack of multilingual resources for QA tasks, and more generally for widely used {\em leaderboard-oriented} tasks.
\end{comment}

\paragraph{Hidden distillation effect}

% Discussion avec Victor San sur le sujet:
%(https://arxiv.org/abs/2010.03058 & https://arxiv.org/abs/1911.05248) 
%en gros quand tu prunes, tu enleves d'abord les features de "robustness"/generalization
%et apres les features qui permettent de resoudre ton dataset
%j'avais des exps comme ca ou la perf de transfer drop sans que la perf du signal/dataset de training drop, et puis plus tu prunnes, celle du signal de training droppe
%tout ca c'est en grosse partie une question de capacity du modele: plus ton modele est gros, plus t'as de la place pour mettre de la generalization dans ton modele
%(roberta-large transfere bien mieux que roberta-base meme si c'est le meme pre-training)

The relative improvement for our best synthetic configuration \emph{+synth-trans}, over the baseline, is above 60\% EM for MiniLM (from 29.5 to 49.5 on XQuAD and from 26.0 to 41.4 on MLQA). 
Significantly higher than that observed for XLM-R (+11.7\% on XQuAD and +2.71\% on MLQA), it indicates that XLM-R provides superior cross-lingual transfer abilities than MiniLM, a fact that we hypothesize due to distillation.
% It is way more important than for XLM-R, where the EM improved by +11.7\% on XQuAD and +2.71\% on MLQA.
% It indicates that XLM-R contains more cross-lingual transfer abilities than MiniLM.
% Since the latter is a distilled version of the former, we hypothesize that such abilities may have been lost during the distillation process. 
Such loss of generalisation can be difficult to identify, and opens questions for future work.

\iffalse

\paragraph{QA, a still challenging task for lower resource languages}
Factoid QA task has been criticized for being too easy and being easily solvable. Most of the times the answer can be easily identified, given simple heuristics: e.g. a ``When'' question is answered by a date. SQuAD-v2 \cite{rajpurkar2018know} was introduced to add more challenges with unanswerable questions. The research community is now building non-factoids based datasets. 

The motivation for this paper, is to cope for the lack of training data. Therefore, critics regarding the simplicity of the task are akin to a first-world problem, from a lower resource languages point of view so to speak. Our work is important as it allows to quickly alleviate the lack of training data, and thus allows us to focus on other striking issues such as domain adaptation and robustness.
\else
\paragraph{QA, an unsolved task for lower resource languages}
% someone is here ?  I wnated to to make a pass here  (djamé)

Factoid QA tasks have been criticized for being a too easy task: the answer can often be identified given simple heuristics: e.g. a ``When'' question is answered by one of the ``date" spans in the context \cite{kocisky-etal-2018-narrativeqa, kwiatkowski-etal-2019-natural}. SQuAD-v2 was for instance introduced to increase the difficulty of the task by adding unanswerable questions. The research community is now moving towards the construction of long context questions and non-factoid QA datasets \cite{dulceanu-etal-2018-photoshopquia,hashemi-etal-2019-Antique,fan-etal-2019-eli5,lewis2020retrieval}. 
In any case, the motivation of this work was to cope for the lack of training data for under-served languages in the QA domain which was severely impacting models performance. Therefore, potential criticisms regarding the simplicity of the task do not apply if seen from a lower-resource language scenario: our work deals with alleviating the lack of native training data, allowing us to focus our future work on further important issues such as domain adaptation, robustness and explainability in low-resource contexts.

\fi

%In contemporaneous work, \citet{shakeri2020multilingual} adress with a similar approach multilingual QA. We detail the differences in our discussion, section \ref{discussion}.

\section{Conclusion}

In this work, we presented a method to generate synthetic QA dataset in a multilingual fashion, showing how QA models can benefit from it and reporting large improvements over the baselines. The proposed approach contributes to fill the gap between English and other languages, and is shown to generalize for languages not present in the synthetic corpus (e.g.~French, Italian, Korean). 

In future work, we plan to investigate whether the proposed data augmentation method could be applied to other multilingual tasks, such  as classification.
%We will also investigate more in depth about how one can control the generation language of a model, and extrapolate on unseen ones. 
We will also experiment more in depth with different strategies to control the target language of a model, and extrapolate on unseen ones. 
%We believe that this difficulty to extrapolate might highlight an important limitation of current language models.

\section{Acknowledgments}
        Djamé Seddah was partly funded by the French Research National Agency via the ANR project ParSiTi (\mbox{ANR-16-CE33-0021}), Arij Riabi was partly funded by Benoît Sagot's chair in the PRAIRIE institute  as part of the French national agency ANR ``Investissements d’avenir'' programme (\mbox{ANR-19-P3IA-0001}) and by the Counter H2020  European project (grant 101021607).
        
% CONTROL EXP CROSS REPRESENTATION

\bibliography{anthology,emnlp}
\bibliographystyle{acl_natbib}

\clearpage

\appendix

\section*{Appendix}
\renewcommand{\thesubsection}{\Alph{subsection}}

\subsection{On target language control for text generation}
When relying on the translated versions of SQuAD, the target language for generating synthetic questions can easily be controlled, and results in fluent and relevant questions in the different languages%(see Table~\ref{tab:examples_generated_questions})
. 
However, one limitation of this approach is that synthetic questions can only be generated in the languages that were available during training: the \texttt{<LANG>} prompts are special tokens that are randomly initialised when fine-tuning QG on SQuAD$_{trans;qg}$: before fine-tuning, they bear no semantic relation with the corresponding language names (``English'', ``Español'' etc.), thus the learned representations for the \texttt{<LANG>} tokens are limited to the languages present in the training set.
% They are not semantically related, before fine-tuning, to the name of corresponding language (``English'', ``Español'' etc.) and, thus, the learned representation for the \texttt{<LANG>} tokens is limited to the languages present in the training set.

To the best of our knowledge, no method allows so far to generalize this target control to an unseen language. It would be valuable, for instance, to be able to generate synthetic data in Korean, French and Italian, without having to translate the entire SQuAD$-en$ dataset in these three languages to then fine-tune the QG model.

To this purpose, we report -- alas, as a negative result -- the following attempt: 
instead of controlling the target language with a special, randomly initialised, token, we used a token  semantically related to the language-word: ``English'', ``Español'' for Spanish, or ``中文'' for Chinese. 
The intuition is that the model might adopt the correct language at inference, even for a target language unseen during training.\footnote{With \emph{unseen during training}, we mean \emph{not present in the QG dataset}; obviously, the language should have been present in the first self-supervised stage.} 
A similar intuition has been explored in GPT-2: the authors report an improvement for summarization when the input text is followed by ``TL;DR" (i.e. Too Long Didn't Read).

At inference time, we evaluated this approach on French with the prompt \texttt{language=Français}. Unfortunately, the model did not succeed to generate text in French. 
Controlling the target language in the context of multilingual text generation remains under-explored, and progress in this direction could have direct applications to improve this work, and beyond.

\begin{table}[!h]
    \centering \small \footnotesize
    \begin{tabular}{llllllll} \toprule
q/c & en   & es  & de  & ar  & hi  & vi   & zh  \\  \midrule
en & 14.5  & 8.9 & 7.2 & 5.9 & 6.5 & 8.4  & 6.0   \\
es &  9.0  & 10. & 6.6 & 4.2 & 5.9 & 6.3  & 4.6   \\
de &  6.2  & 4.8 & 6.3 & 3.1 & 3.7 & 5.0  & 3.2   \\
ar &  2.8  & 2.2 & 2.4 & 3.3 & 2.0 & 2.3  & 2.1   \\
hi &  7.9  & 6.7 & 6.6 & 5.8 & 8.3 & 6.6  & 5.2   \\
vi &  9.1  & 7.3 & 7.2 & 6.0 & 6.5 & 12.3 & 6.1   \\
zh &  9.2  & 8.0 & 7.8 & 6.1 & 7.2 & 8.0  & 15.0  \\ \bottomrule
\end{tabular}

    \caption{BLEU4 scores on MLQA test for QG\textsubscript{synth+trans} model.}
    \label{tab:mat_QG_MLQA_synth+trans}
\end{table}

\begin{table}[!h]
    \centering \small \footnotesize
    \begin{tabular}{llllllll} \toprule
q/c & en & es & de & ar & hi & vi & zh \\ \midrule
en &  21.7 & 4.73 & 4.58 & 2.47 & 2.58 & 3.02 & 3.11  \\
es &   0.8  & 1.23 & 0.38 & 0.0  & 0.0  & 0.22 & 0.11\\
de &  1.4  & 0.85 & 1.32 & 0.0  & 0.0  & 0.22 & 0.0   \\
ar &  0.0  & 0.0  & 0.0  & 0.0  & 0.0  & 0.0  & 0.0   \\
hi &   0.0  & 0.21 & 0.0  & 0.0  & 0.0  & 0.0  & 0.0    \\
vi &   0.55 & 0.5  & 0.25 & 0.0  & 0.0  & 0.34 & 0.0   \\
zh &   0.0  & 0.0  & 0.0  & 0.0  & 0.0  & 0.0  & 0.0  \\ \bottomrule
\end{tabular}
    \caption{BLEU-4 scores on MLQA test for QG\textsubscript{synth} model.}
    \label{tab:mat_MLQA_synth}
\end{table}

\begin{table}[!h]
    \centering \small \footnotesize
    \begin{tabular}{llllllll} \toprule
q/c & en & es & de & ar & hi & vi & zh \\ \midrule
en & 83.9 & 74.8 & 70.1 & 67.8 & 72.3 & 74.1 & 69.3  \\
es & 78.0 & 74.3 & 68.1 & 63.6 & 67.4 & 67.6 & 69.1  \\
de & 75.6 & 72.9 & 70.1 & 65.0 & 67.5 & 68.5 & 70.4   \\
ar & 61.3 & 58.4 & 56.8 & 66.7 & 57.7 & 56.5 & 69.8  \\
hi & 70.9 & 62.1 & 58.9 & 56.8 & 70.9 & 61.3 & 69.2  \\
vi & 71.0 & 64.1 & 60.5 & 59.7 & 63.7 & 74.7 & 69.9   \\
zh & 67.1 & 62.5 & 59.0 & 56.7 & 60.7 & 63.2 & 69.3  \\ \bottomrule
\end{tabular}
    \caption{F1 score on MLQA for XLM-R model finetuned on SQuAD\textsubscript{en}.}
    \label{tab:mat_MLQA_squad}
\end{table}

\begin{table}[!h]
    \centering \small \footnotesize
    \begin{tabular}{llllllll} \toprule
q/c & en & es & de & ar & hi & vi & zh \\ \midrule
en & 83.9 & 75.4 & 70.9 & 68.9 & 72.8 & 75.6 & 66.8   \\
es & 81.0 & 74.4 & 71.8 & 66.6 & 70.2 & 72.5 & 65.7   \\
de & 81.4 & 75.2 & 70.8 & 69.3 & 70.2 & 74.4 & 65.1  \\
ar & 76.3 & 68.9 & 67.3 & 66.6 & 65.4 & 70.4 & 61.7   \\
hi & 78.5 & 70.5 & 64.5 & 63.6 & 70.8 & 71.1 & 63.2   \\
vi & 78.0 & 72.0 & 66.4 & 65.2 & 68.5 & 74.7 & 64.5  \\
zh & 77.8 & 70.1 & 67.0 & 64.9 & 68.1 & 71.8 & 67.7  \\ \bottomrule
\end{tabular}
    \caption{F1 score  on MLQA for XLM-R\textsubscript{+synth-trans} model.}
    \label{tab:mat_MLQA_gen}
\end{table}

\begin{table*}[!h]
    \centering \footnotesize
    \begin{tabular}{lccccccccccc} \toprule
q/c & en & es & de & ar & hi & vi & zh & ru & th & tr & el\\ \midrule
en & 87.4 & 82.0 & 80.7 & 75.6 & 79.0 & 78.1 & 73.0 & 79.0 & 73.2 & 74.8 & 79.9 \\
es & 80.6 & 84.2 & 76.2 & 71.3 & 72.7 & 72.8 & 67.5 & 76.4 & 69.6 & 70.6 & 75.8 \\
de & 79.8 & 76.4 & 83.5 & 71.0 & 72.9 & 72.4 & 68.1 & 75.6 & 68.3 & 71.4 & 75.5 \\
ar & 65.3 & 63.1 & 62.0 & 77.8 & 64.1 & 61.9 & 58.8 & 63.9 & 62.8 & 57.4 & 65.2\\
hi & 72.9 & 64.8 & 65.7 & 63.7 & 75.8 & 64.0 & 61.8 & 66.9 & 63.5 & 59.6 & 67.2\\
vi & 74.5 & 69.8 & 70.7 & 67.7 & 67.7 & 80.6 & 66.9 & 70.5 & 66.6 & 64.8 & 70.6 \\
zh & 70.7 & 65.8 & 64.1 & 64.1 & 65.6 & 67.4 & 81.8 & 67.4 & 64.8 & 60.8 & 66.9 \\ 
ru &79.8 & 78.0 & 76.7 & 70.2 & 72.5 & 74.6 & 67.3 & 81.1 & 69.6 & 70.7 & 75.6\\
th &49.6 & 42.1 & 44.1 & 49.8 & 51.9 & 49.0 & 53.9 & 50.4 & 74.9 & 37.1 & 48.8\\
tr &72.6 & 64.9 & 68.5 & 65.5 & 68.1 & 64.1 & 62.4 & 70.3 & 64.3 & 76.7 & 71.3 \\
el &70.8 & 69.1 & 69.3 & 63.3 & 65.6 & 65.0 & 63.0 & 70.8 & 64.2 & 62.3 & 81.3  \\\bottomrule
\end{tabular}
\caption{F1 score on XQuAD for XLM-R model finetuned on SQuAD\textsubscript{en}.}
\label{tab:mat_xquad_squad}

\end{table*}
\begin{table*}[!h]
    \centering \footnotesize
  \begin{tabular}{lccccccccccc} \toprule
q/c & en & es & de & ar & hi & vi & zh & ru & th & tr & el\\ \midrule
en & 87.0 & 82.9 & 81.6 & 77.8 & 81.5 & 81.0 & 77.5 & 80.9 & 77.7 & 74.8 & 81.4\\
es & 83.5 & 84.1 & 79.2 & 74.9 & 78.0 & 78.7 & 76.5 & 78.9 & 76.3 & 72.7 & 78.8\\
de & 82.8 & 80.3 & 83.1 & 75.1 & 77.3 & 78.3 & 75.1 & 78.4 & 75.7 & 72.9 & 78.3\\
ar & 78.4 & 77.2 & 75.6 & 77.5 & 72.8 & 74.3 & 72.1 & 74.2 & 72.5 & 69.5 & 74.2 \\
hi & 80.6 & 77.5 & 76.4 & 71.6 & 77.5 & 75.2 & 73.9 & 76.3 & 73.6 & 70.2 & 75.9\\
vi & 81.3 & 79.6 & 77.5 & 73.0 & 75.6 & 80.7 & 74.7 & 76.5 & 74.5 & 71.8 & 76.0\\
zh & 80.2 & 78.3 & 76.9 & 72.1 & 74.3 & 76.1 & 85.1 & 77.0 & 74.1 & 70.1 & 75.7\\ 
ru & 82.2 & 80.4 & 78.8 & 74.1 & 75.9 & 78.4 & 76.5 & 80.8 & 75.4 & 72.9 & 78.6\\
th & 79.1 & 76.7 & 75.1 & 71.0 & 73.5 & 75.1 & 73.6 & 75.3 & 77.2 & 68.8 & 74.1 \\
tr & 80.2 & 77.7 & 76.4 & 71.9 & 74.6 & 74.2 & 74.7 & 77.8 & 73.3 & 75.2 & 75.6 \\
el & 82.0 & 80.1 & 78.5 & 73.8 & 77.3 & 77.3 & 75.4 & 78.6 & 75.3 & 72.2 & 80.1 \\\bottomrule
\end{tabular}
\caption{F1 score  on XQuAD for XLM-R\textsubscript{+synth-trans} model.}
\label{tab:mat_xquad_f1}
\end{table*}

\subsection{Question Generation Scores}
We report the BLEU-4 scores for MLQA on QG\textsubscript{synth+trans} on Table \ref{tab:mat_QG_MLQA_synth+trans} and QG\textsubscript{synth} on Table \ref{tab:mat_MLQA_synth}. In addition, we report the F1 scores for XLM-R finetuned on SQuAD\textsubscript{en} and XLM-R\textsubscript{+synth-trans} on all the language pairs, on both MLQA and XQuAD in Tables \ref{tab:mat_MLQA_squad}, \ref{tab:mat_MLQA_gen}, \ref{tab:mat_xquad_squad} and \ref{tab:mat_xquad_f1}.
 
%\newpage

\subsection{Qualitative Evaluation}

We report in Tables \ref{tab:human_eval_para_1}, \ref{tab:human_eval_para_2}, and \ref{tab:human_eval_para_3} different examples that we analysed in our manual qualitative analysis, discussed at the end of section 5.1 in the main paper.

\begin{table*}
    \centering \footnotesize
    \begin{tabular}{p{.95\linewidth}}
    \toprule
    
        \textbf{Paragraph}
       Kochi was the centre of Indian spice trade for many centuries, and was known to the Yavanas (Greeks and Romans) as well as Jews, Syrians, Arabs, and Chinese since ancient times. It rose to significance as a trading centre after the port Muziris around Kodungallur (Cranganore) was destroyed by \textbf{massive flooding of Periyar} in 1341. The earliest documented references to Kochi occur in books written by Chinese voyager \textbf{Ma Huan} during his visit to Kochi in the 15th century as part of Admiral Zheng He's treasure fleet. There are also references to Kochi in accounts written by Italian traveller Niccolò Da Conti, who visited Kochi in 1440.
\\
        \midrule
               
        \textbf{Answer} massive flooding of Periyar\\
        \textbf{QG\textsubscript{synth+trans} (target language = de)} \textbf{Welche Veranstaltung hat den Angriff auf Kochi im Jahr 1341 verursacht?} (tr: \textit{Which event
caused the attack on Kochi in 1341?})
        \\
        \textbf{QG\textsubscript{synth+trans} (target language = ar)} \AR{
        ما هو الحدث الذي تسبب في تدمير كاتماندو? } (tr: \textit{What event caused the destruction of Kathmandu?})
        \\

        \midrule
       
        \textbf{Answer} Ma Huan \\
        \textbf{QG\textsubscript{synth+trans} (target language = de)} \textbf{Welcher chinesische traveller hat die frühesten Erinnerungen an Kochi geschrieben?} (tr: \textit{Which Chinese traveler wrote the earliest memories of Kochi?})\\
        \textbf{QG\textsubscript{synth+trans} (target language = ar)} \AR{?}15\AR{
من كتب اول اشارة الى كوتي في القرن ال } (tr: \textit{Who wrote the first reference to Coty in the 15th century?})\\
        
        \bottomrule
    \end{tabular}
    \caption{Paragraph 1 and different questions analysed during out human evaluation.}
    \label{tab:human_eval_para_1}
\end{table*}
\begin{table*}
    \centering\footnotesize
\begin{tabular}{p{.95\linewidth}}
    \toprule
    
        \textbf{Paragraph} Portuguese navigator, \textbf{Pedro Álvares Cabral} founded the first European settlement in India at Kochi in 1500. From 1503 to 1663, Fort Kochi (Fort Emmanuel) was ruled by Portugal. This Portuguese period was a harrowing time for the Saint Thomas Christians and the Jews, as the Inquisition was active in Portuguese India. Kochi hosted the grave of Vasco da Gama, the first European explorer to set sail for India, who was buried at \textbf{St. Francis Church} until his remains were returned to Portugal in 1539. The Portuguese rule was followed by that of the Dutch who renamed Fort Immanuel as Fort Stormsburg. In meantime, the Royal Family of Kochi relocated the capital of Kochi Kingdom to Thrissur, leaving nominal authority over Islands of Kochi. In 1664, Fort Kochi Municipality was established by Dutch, making it the first municipality in Indian subcontinent, which got dissolved when Dutch authority got weaker in the 18th century. The remaining part of Kochi were governed by governors of Kochi Kingdom. By 1773, the Mysore ruler \textbf{Hyder Ali} extended his conquest in the Malabar region to Kochi forcing it to become a tributary of Mysore. The hereditary Prime Ministership of Kochi held by the Paliath Achans ended during this period.
\\
        \midrule
               
        \textbf{Answer} Pedro Álvares Cabral\\
        \textbf{QG\textsubscript{synth+trans} (target language = de)} \textbf{Wer hat die erste europäische Siedlung in Kochi gegründet?} (tr: \textit{Who founded the first European settlement in Kochi?})
        \\
        \textbf{QG\textsubscript{synth+trans} (target language = ar)} \AR{?}\textbf{1500} \AR{من الذي اسس اول تسوية الاوروبية في الهند في عام } (tr: \textit{Who founded the first European settlement in India in the year 1500?})
        \\

        \midrule
        \textbf{Answer} St. Francis Church\\
        \textbf{QG\textsubscript{synth+trans} (target language = de)} \textbf{Wo wurde Vasco da Gama begraben?} (tr: \textit{Where was Vasco da Gama buried?})
        \\
        \textbf{QG\textsubscript{synth+trans} (target language = ar)} \AR{?} \textbf{da Gama} \AR{
اين تم دفن فازاو} (tr: \textit{Where was Vazaw da Gama buried?})
        \\
        \midrule
        \textbf{Answer} Hyder Ali\\
        \textbf{QG\textsubscript{synth+trans} (target language = de)} \textbf{Wer war der Herrscher von Mysore im Jahr 1773?} (tr: \textit{Who was the ruler of Mysore in 1773?})
        \\
        \textbf{QG\textsubscript{synth+trans} (target language = ar)} \AR{?}\textbf{1773} \AR{من كان الحاكم في ميسور في عام } (tr: \textit{Who was the ruler of Mysore in 1773?})
        \\
        
        \bottomrule
    \end{tabular}
    \caption{Paragraph 2 and different questions analysed during out human evaluation.}
    \label{tab:human_eval_para_2}

\end{table*}
\begin{table*}
    \centering\footnotesize
    \begin{tabular}{p{.95\linewidth}}
    \toprule
    
        \textbf{Paragraph} 
Taiwan is an island country in East Asia. The main island, known historically as Formosa, makes up \textbf{99\% }of the area controlled by the ROC, measuring \textbf{35,808 square kilometres }(13,826 sq mi) and lying some \textbf{180 kilometres} (112 mi) across the Taiwan Strait from the southeastern coast of mainland China. The East China Sea lies to its north, the Philippine Sea to its east, the \textbf{Luzon Strait} directly to its south and the \textbf{South China Sea} to its southwest. Smaller islands include a number in the Taiwan Strait including the Penghu archipelago, the Kinmen and Matsu Islands near the Chinese coast, and some of the South China Sea Islands.
\\
        \midrule
               
        \textbf{Answer} 99\%
 \\
        \textbf{QG\textsubscript{synth+trans} (target language = de)} \textbf{Welcher Prozentsatz der Gebiete von Taiwan wird von der ROK kontrolliert?} (tr: \textit{What percentage of the areas of Taiwan is controlled by the ROK?})
        \\
        \textbf{QG\textsubscript{synth+trans} (target language = ar)} \AR{ما هي نسبة الاراضي التي تسيطر عليها ال (ديوان)?} (tr: \textit{What is the percentage of lands controlled by the (Divan)?})
        \\

        \midrule
        \textbf{Answer} 35,808 square kilometres\\
        \textbf{QG\textsubscript{synth+trans} (target language = de)} \textbf{Wie groß ist die RAF?} (tr: \textit{How big is the RAF?})
        \\
        \textbf{QG\textsubscript{synth+trans} (target language = ar)} \AR{
كم ميل مربع هي الجزيرة?} (tr: \textit{How many square miles is the island?})
        \\

        \midrule
        \textbf{Answer} Luzon Strait\\
        \textbf{QG\textsubscript{synth+trans} (target language = de)} \textbf{Welcher Fluss ist direkt zum Süden von Taiwan?} (tr: \textit{Which river is directly to the south of Taiwan?})
        \\
        \textbf{QG\textsubscript{synth+trans} (target language = ar)} \AR{
ما هو اسم السكك الحديدية التي تقع بالقرب منها?} (tr: \textit{What is the name of the nearby railway?})
        \\
        \midrule
       \textbf{Answer} South China Sea\\
        \textbf{QG\textsubscript{synth+trans} (target language = de)} \textbf{Welches Meer ist im Südwesten von Taiwan?} (tr: \textit{Which sea is in southwest of Taiwan?})
        \\
        \textbf{QG\textsubscript{synth+trans} (target language = ar)} \AR{
ما هو المحيط الذي ينظر اليه على الجانب الغربي من تايوان?} (tr: \textit{What is the ocean seen on the western side of Taiwan?})
        \\
        \midrule
        \textbf{Answer} 180 kilometres\\
        \textbf{QG\textsubscript{synth+trans} (target language = de)} \textbf{Wie weit ist die RAF von Taiwan aus der südlichen Küste von China?} (tr: \textit{How far is the RAF from Taiwan from the southern coast of China?})\\
        \textbf{QG\textsubscript{synth+trans} (target language = ar)} \AR{
كم من الوقت تبعد جزر تايوان عن ساحل الصين?} (tr: \textit{How long are the Taiwan Islands from the coast of China?})\\
        
        \bottomrule
    \end{tabular}
    \caption{Paragraph 3 and different questions analysed during out human evaluation.}
    \label{tab:human_eval_para_3}
\end{table*}

\clearpage

\begin{figure*}[h!]
\centering
\begin{minipage}{0.49\linewidth}
\centering
\includegraphics[width=\linewidth]{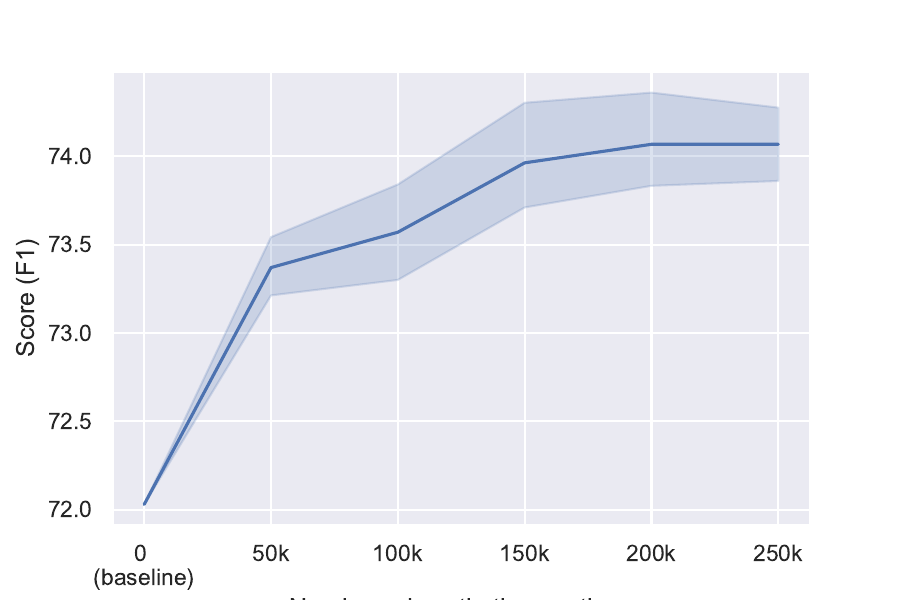}
  \caption{SQuAD-it}
  \label{fig:squad-it}
\end{minipage}\hfill
\begin{minipage}{0.49\linewidth}
\includegraphics[width=\linewidth]{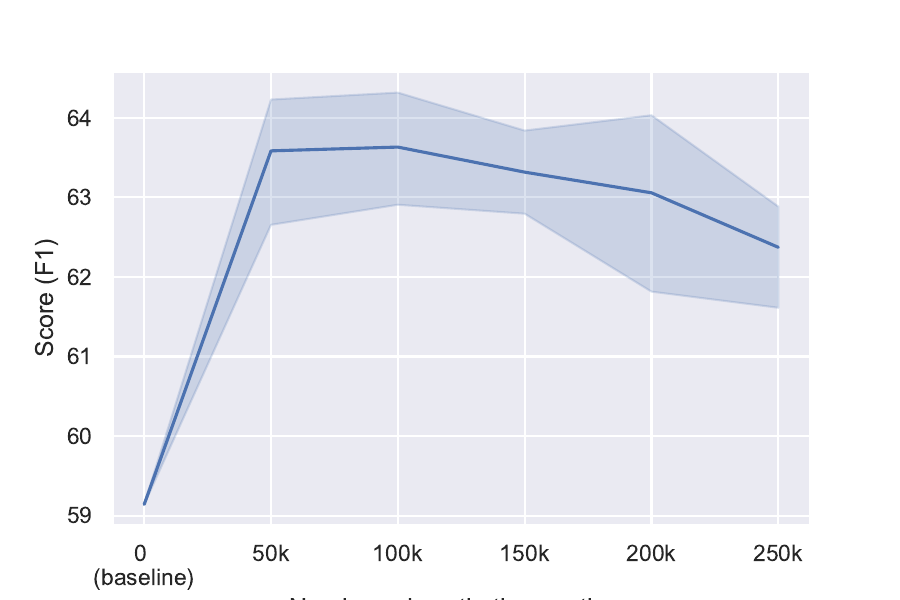}
  \caption{PIAF (fr)}
  \label{fig:piaf}
\end{minipage}\par
\vskip\floatsep% normal separation between figures
\includegraphics[width=0.49\linewidth]{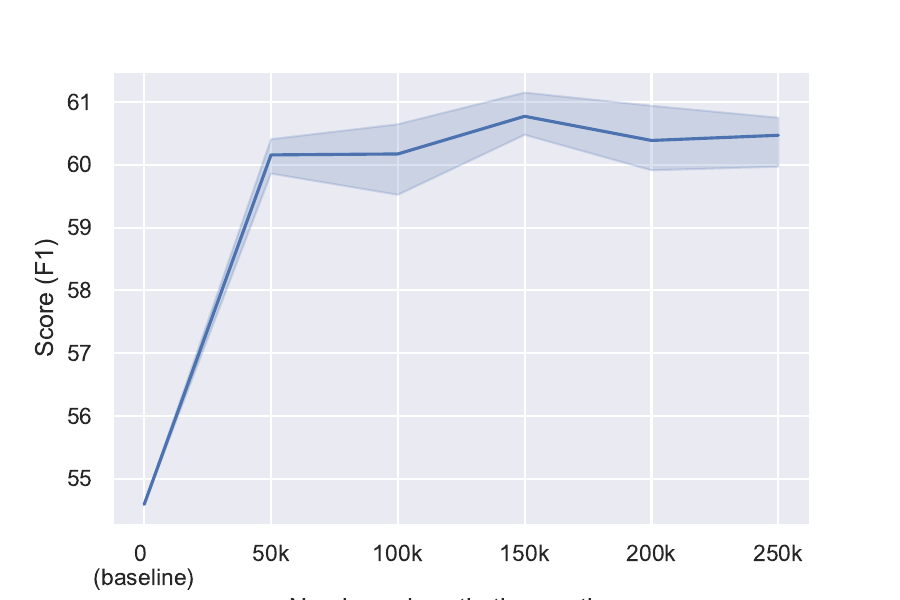}
  \caption{KorQuAD}
  \label{fig:korquad}
  \vspace{1em}
  \begin{flushleft}
 \subsection{Learning Curves for Unseen Languages}
We show on Figures \ref{fig:squad-it},\ref{fig:piaf},\ref{fig:korquad} the results of three learning curves for respectively  SQuAD-it, PIAF (fr) and KorQuad where the models are trained on different amount of synthetic questions in our \emph{All Languages} setting.

The synthetic questions are sampled among all the five languages in MLQA . The standard deviation over four different seeds for the sampling are displayed through the confidence interval (light blue) around the averaged main curves.
We observe  that for SQuAD-it and KorQuAD  the performances increase significantly at the beginning, then remain mostly stable, while for PIAF (fr) the best performances are obtained with 100k of additional synthetic data, a slight improvement from 50k additional questions before starting to decrease. 
 
  \end{flushleft}

\end{figure*}

\end{CJK*}
\end{document}